\documentclass[onecolumn,10pt]{asme2ej}

\usepackage{epsfig}
\usepackage{graphicx}
\usepackage{tabularx}
\usepackage{caption}
\usepackage{array}
\usepackage{subcaption}
\usepackage{setspace}
\usepackage{fancyhdr}
\title{Literature Review on Endoscopic Robotic Systems in Ear and Sinus Surgery}

\author{Guillaume Michel \thanks{Address all correspondence to this author.}
    \affiliation{
	Surgeon, ENT department,
	CHU de Nantes\\
	1, place A. Ricordeau,	44093 Nantes, France\\
    Email: guillaume.michel@chu-nantes.fr
    }
}
\author{Durgesh Haribhau Salunkhe 
    \affiliation{ PhD student,
	 Laboratoire des Sciences du Num\'erique de Nantes\\
       UMR CNRS 6004, 1 rue de la No\"e,
        Nantes, France\\
        Email: durgesh.salunkhe@ls2n.fr
    }
}
\author{Philippe Bordure 
    \affiliation{ Professor, ENT department,
	CHU de Nantes\\
	1, place A. Ricordeau\\
	44093 Nantes, France\\
    Email: philippe.bordure@chu-nantes.fr
    }
}
\author{Damien Chablat 
    \affiliation{CNRS Senior researcher \\
        Laboratoire des Sciences du Num\'erique de Nantes\\
        UMR CNRS 6004, 1 rue de la No\"e,
        44321 Nantes, France\\
        Email: damien.chablat@cnrs.fr
    }
}

\begin{document}
\doublespacing
\maketitle

\begin{abstract}
{\it In otolaryngologic surgery, endoscopy is increasingly used to provide a better view of hard-to-reach areas and to promote minimally invasive surgery. However, the need to manipulate the endoscope limits the surgeon's ability to operate with only one instrument at a time. Currently, several robotic systems are being developed, demonstrating the value of robotic assistance in microsurgery. The aim of this literature review is to present and classify current robotic systems that are used for otological and endonasal applications. For these solutions, an analysis of the functionalities in relation to the surgeon's needs will be carried out in order to produce a set of specifications for the creation of new robots.}
\end{abstract}

\begin{nomenclature}
\entry{DOF}{Degree Of Freedom}
\entry{RCM}{Remote Center of Motion}
\end{nomenclature}
\section{Introduction}
\label{Introduction}
In medicine, surgical robots play an important role in revolutionizing  conventional surgery procedures. In otologic and endonasal surgeries, robotic systems could play an important role, because of the challenging anatomy and the need  for the surgeon to handle the endoscope. 

\begin{figure}[ht]
    \centering
        \begin{subfigure}{0.45\columnwidth}
        	\centering
        	\includegraphics[width=\columnwidth]{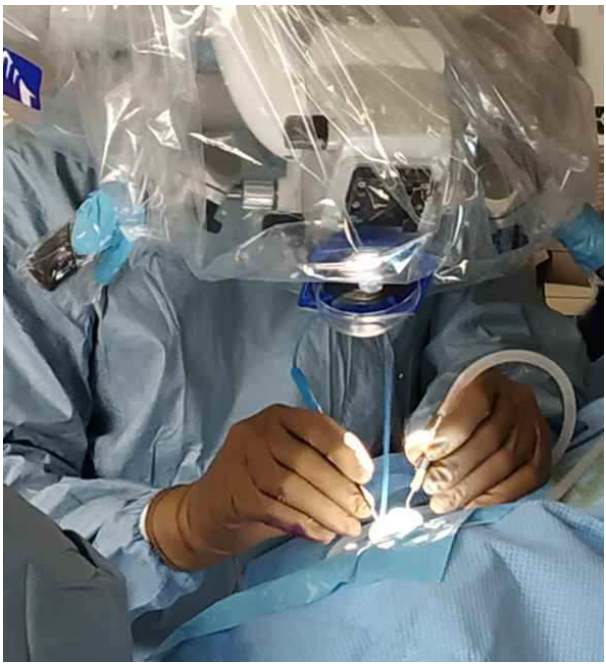}
        	\caption{Surgeon using 2 instruments with a microscope}
        	\label{fig:micro_2hand}
        \end{subfigure}
        ~
    	\begin{subfigure}{0.5\columnwidth}
    		\centering
    		\includegraphics[width=\columnwidth]{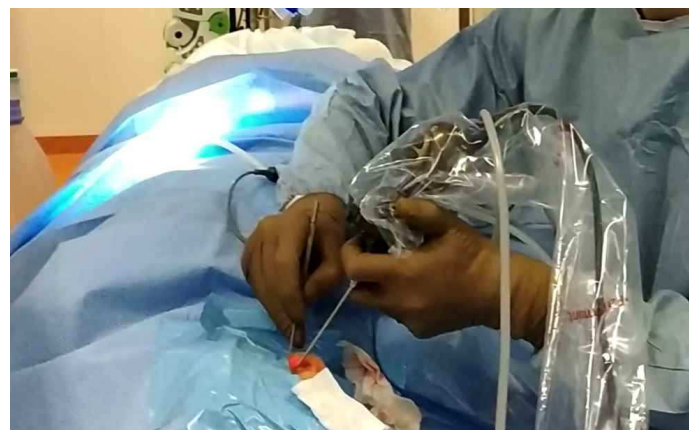}
    		\caption{Use of 1 hand to hold endoscope limits the instruments}
    		\label{fig:endo_1hand}
    	\end{subfigure}
    	\caption{The comparison of the number of instruments possible to use simultaneously while using an endoscope and a microscope.}
    	\label{fig:endo_micro_hand}
    \end{figure}

The middle ear is an anatomical entity of small volume with multiple fragile elements not to be damaged. Operations are traditionally performed under binocular magnifiers, which allows the surgeon to use both hands for a micro-instrument and a suction tool, as illustrated in Fig. \ref{fig:micro_2hand}. 
But microscopic vision is limited by the external ear canal and the surgeon often has to sacrifice bone to access certain areas of interest.  The development of endoscopic otologic surgery allows better vision of hard-to-reach areas, without bone sacrifice. Endoscopic cholesteatoma surgery has demonstrated its advantage to detect recurrences \cite{ohki_residual_2017}, less post-operative pain and a shorter healing time than with microscopic procedures~\cite{magliulo_endoscopic_2018}.

But the surgeon can only use one tool at a time in endoscopic surgery in contrast to two tools with the use of a microscope as illustrated in Fig.~\ref{fig:endo_1hand} \cite{chablat2021workspace}.
  
This makes endoscopic surgeries cumbersome, as the surgeon must switch from one tool to the other to operate and manage bleeding in the ear. The inability to use both hands for the operation leads to frustration and fatigue for the surgeon. Using a robotic arm to manipulate the endoscope as needed can dramatically improve the performance of otologic surgeries.

Contrary to ear surgery, endonasal surgery with endoscopes  has been the reference for almost thirty years. But, as in otologic surgery, one hand still holds the endoscope. The analysis of senior surgeons' gestures  \cite{rapport_SFORL} shows that 20 to 50 \% of the total time of surgery is spent sucking up the accumulated blood to allow the visualisation of the operating field with water, or trying to find the instrument of adequate shape. Another difficulty is the impossibility of distracting tissue before cutting with one hand \cite{rapport_SFORL}. These limitations could be overcome by allowing the surgeon to use both hands rather than working single handed. 

The objective of this review is to present robotic system studied for otologic and endonasal surgery, and to highlight those designed for endoscopic solutions. 

\section{Method} 

A literature search was carried out using the PubMed, MedLine and Google Scholar databases.
The keywords used in this literature search included: ear surgery, temporal bone, robotics, cochlear implant, mastoid surgery, sinus surgery, endonasal, endoscopic surgery, and appropriate combinations of these terms. 
In order to ensure the most comprehensive search possible, there were no exclusions based on date of publication. 
After the initial search, articles were selected if the robot studied was specifically designed for ear or sinus surgery. 
Robots designed for neurosurgery were excluded, even if the surgical approach was the nasal cavity, due to the difference in workspace (narrower and deeper) and user. 

\section{Otological robotic systems}

Many robotic systems have been studied in otologic surgery, because surgeon faces many challenges: gestures are meticulous, anatomical structures are microscopic and fragile, the field of vision is limited. 
Robotic system are designed to help the surgeon in these difficulties. 
According to their medical application, otological robotic systems can be classified in three groups: 
\begin{enumerate}
    \item drilling robots, designed for drilling the mastoid bone during cochlear implantation surgery. The objective is to fasten and secure the gesture in this complex anatomic area; 
    \item  microscopic gesture robots, studied to perform precise procedures for the surgeon, like ossicular replacement or cochlear implantation. The objective is to avoid tremors and gain precision;
    \item endoscopic holding robots, designed to hold the endoscope and free a hand from the surgeon. The objective is to let the surgeon operate with both hands, as the robot hold the endoscope. 
\end{enumerate}
\subsection{Drilling robots}
\label{Drilling robots}
The majority of otological robotic systems are dedicated to bone drilling for the cochlear implants. They share the ability to drill according to a predefined trajectory, thanks to a preoperative scanner. These robots drill under the supervision of the surgeon, in order to secure the drilling close to the facial nerve.

At a pre-clinical level, a robotic system was proposed \cite{hussong_conception_2008, majdani_robot-guided_2009} to drill a tunnel using an industrial robot (KUKA KR3). Another team \cite{danilchenko_robotic_2011} used an industrial robot (Mitsubishi RV-3S) for bone drilling, with an optical tracking system to secure the tool positioning. This system, called OTOBOT, was tested on cadaveric study. But precision was low and, as for most of industrial robots, not compatible with a clinical application. 

Dillon et al. \cite{dillon_compact_2015} proposed a compact robot attached to the mastoid bone. This system has the advantage of eliminating the needs to track the patient's head. But the accuracy again seems rather low (0.5 mm). Microtable \cite{kratchman_design_2011} is another system anchored to the patient's head. The parallel architecture based on the Gough-Stewart platform offers a good targeting error but has only been tested on petrous bone at a preclinical level. Kobler et al. \cite{kobler} used the same architecture to propose a system anchored in the temporal bone. This robot named RoboJig was tested on cadaveric bone for bone drilling and then for insertion of a cochlear implant \cite{kobler_insertion}.

But these robots cannot separate bone from soft tissue. A team from the University of Birmingham \cite{coulson_drill} has studied a drilling system that can make this distinction. The robot analyses the force applied to the tool and the torque of the drill. This system, called ``Smart Micro-Drill'', is able to detect the passage of the drill from bone to soft tissue. Like previous robots, the drilling is carried out automatically, but as the drill moves out of the bone structure, the force decreases to protect the structures of the inner ear, for example. A clinical trial on three patients \cite{coulson_trial} proved that cochleostomy was safe and preserved the membranous labyrinth. 

Another robotic platform has been designed at the ARTORG centre \cite{bell_self-developed_2012} and reduces targeting error by using a customised robot adapted to clinical applications.
This 5 DOF serial robot is attached to the operating table.
This study led to a clinical study on nine patients in 2019 \cite{caversaccio_robotic_2019}. The robotic system drilled down to the level of the facial cavity, with intraoperative imaging to control the distance between the drilling system and the anatomy concerned. The surgeon then manually inserts the electrodes. A commercial version of this system (HEARO) is now being developed by CAScination and Med-El, with CE marking since May 2020.
\subsection{Microscopic gesture robots}
\label{Microscopic gesture robots}
These robots are another way of conceiving their contribution to ear surgery. They are designed to replace the surgeon's hand during microscopic gestures, such as the placement of an ossicular prosthesis or the insertion of a cochlear implant. 
Their common objective is to increase the precision of the gesture, by eliminating the surgeon's tremors. 

Entsfellner \cite{inproceedings} developed the MMTS system for stapes surgery. This robot with 6 DOFs and a joystick allows the surgeon to control the tool with an accuracy of approximately 1.6 mm. This precision was not as good as manual precision, as shown in a comparative study with ten surgeons \cite{hofer_mmts}.  

Another system called Steady-Hand \cite{steady_hand} was designed for microsurgery. It is a parallel structure with a RCM and 7 DOF. It was first designed for ophthalmologic surgery and then tested for stapes surgery.Two studies compared robotic surgery with manual stapedo surgery \cite{steady_stapedo, rothbaum}, and the authors concluded that the accuracy is higher with the robotic system than with surgeons of various expertise.

\begin{figure} [ht]
    \begin{subfigure} {0.45\columnwidth}
        \centering
        \includegraphics[width=\columnwidth]{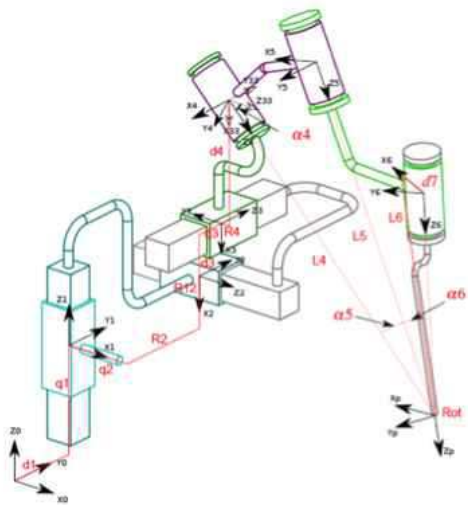}
        \caption{Serial architecture of the Robotol}
        \label{fig:robotol_archi}
    \end{subfigure}
    \begin{subfigure} {0.5\columnwidth}
        \centering
        \includegraphics[width=\columnwidth]{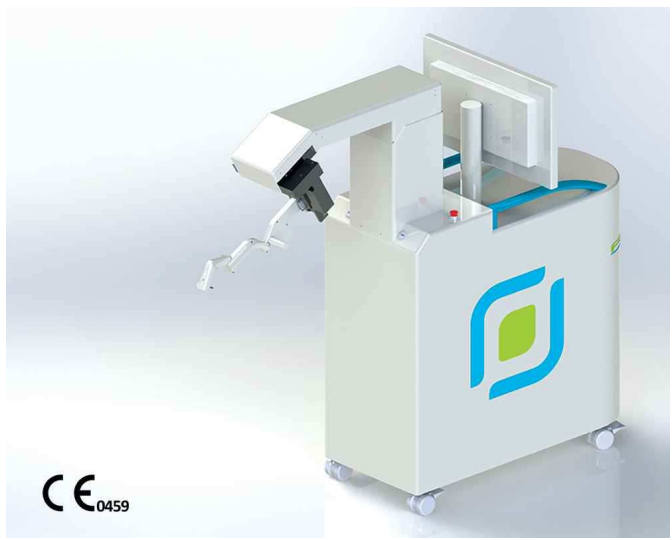}
        \caption{Robotol in its commercialised version with CE marking.}
        \label{fig:robotol_photo}
    \end{subfigure}
    \caption{Robotol system, initially designed for microscopic gesture. Later, a second arm has been added to hold an endoscope. Source: Collin Medical.}
    \label{fig:robotol}
\end{figure}

Then, Robotol was designed by surgeons (Pr Sterkers, Pr N'Guyen, La Pitié Salpêtrière, Paris) and researchers \cite{miroir_robotol:_2010}, and obtained a CE marking.
It is a 7 DOF serial robot, controlled by a SpaceMouse, optimised for ear surgery (Figure \ref{fig:robotol}). The surgeon uses the SpaceMouse to move the end-effector, and to control his gesture under the microscope.Initially developed to replace the surgeon's hand and increase its precision, it could be used to place ossicular prostheses or insert cochlear implants. This robot is the only one of these microscopic gesture robots to be commercialised (Collin Medical) and to have reached a clinical stage. A clinical study \cite{vittoria_robot-based_2020} presented nine stapes surgeries and ten cochlear implant surgeries with the Robotol, without complications. Later in its development, a second arm was developed to hold an endoscope, and will be discussed in the next section. 

\subsection{Endoscopic holding robots}
\label{endos_oreille}
Endoscopic holding robots do not replace the surgeon's gesture, but aim to interact with him. 
Their common objective is to free the surgeon's hand, which usually holds the endoscope. The surgeon can operate with both hands, as under a microscope, but with the advantages of the endoscope presented in Section~\ref{Introduction}. 

A modification of the Robotol \ref{Microscopic gesture robots} was achieved by developing a new arm.
Although its architecture and initial design were conceived for precision gesture, its designers decided to add an endoscope arm \cite{rapport_SFORL}. This arm can be positioned in place of the previous arm and is still controlled by the SpaceMouse.  As the architecture has not changed, the workspace is still optimised for the middle ear. 

Another example is the Exofix, originally designed for abdominal, urological and gynaecological surgery. This system was then studied \cite{endofix_exo} for use in otolaryngological applications. It is a serial robot with 3 DOF with fluidic actuators and controlled by a joystick. However, the study only evaluated endoscopy in the mastoid bone, which is not a common anatomical site for endoscopy. The system is not precise enough for the middle ear and is now being marketed for endonasal application. 

Next, Fichera et al \cite{fichera_through_2017} studied an interesting miniature robotic endoscope which is small enough to pass through the Eustachian tube and allow visualisation of the middle ear. The endoscope itself has a 3 DOF rotatable end, and a chip-tip camera mounted at its end. This system eliminates the need to elevate the tympanic membrane to go inside the middle ear. Another similar system has recently been introduced \cite{comert_2019}. This system offers the doctor the possibility to visually examine the ossicular chain by means of an integrated camera and haptically by means of an integrated tool.  But these devices allow the surgeon to perform a visual examination, or more recently palpation \cite{comert_2019}, but cannot perform surgery. They are therefore devices intended for diagnostic and non-surgical purposes. 
 
Up to now, only the Robotol is suitable for middle ear endoscopy. However, this robot does not eliminate manual control of its mobility, and is not suitable for sinus surgery, due to its initial optimisation, i.e. its working space is too small. All the robotic systems presented are dedicated to ear surgery, but there is no robotic system exclusively dedicated to endoscopic surgery, crossing the border between ear surgery and sinus-nasal surgery.

\section{Endonasal robotic systems}
Several robotic systems have been studied for endonasal surgery. These robots have multiple objectives~\cite{rapport_SFORL}:
\begin{itemize}
     \item[$\bullet$] restore surgery with both hands by releasing the hand that usually holds the endoscope; 
     \item[$\bullet$] follow the surgeon's gesture during the operation. Indeed, several static arms exist  \cite{bras_endos_laryngoscope}, but they remain immobile once locked in a desired position.They cannot (i) adapt to the movement of the instruments for collaborative work, (ii) exit the endoscope when the instruments are extracted from the nostril, and therefore require special endoscopes of extended length \cite{rapport_SFORL};
     \item[$\bullet$] navigate safely in a high-risk anatomical area; the paranasal sinuses are embedded between several important anatomical structures: eye and optic nerve, base of the skull, internal carotid artery.
 \end{itemize}
Endonasal robotic systems may be classified in two groups: those assisting the surgeon by holding the endoscope, and those replacing the surgeon to perform a task. 

\subsection{Endoscopic holding robots}
Endoscopic holding robots are numerous at the preclinical stage, responding to the first objective of releasing the surgeon's hand. But few respond to the other objectives of collaboration and navigation.
They are separate from otological endoscopic holders, because their architecture is too bulky for ear surgery, without consideration for this specific workspace. 

A prototype was presented in 2002 by Koseki et al. \cite{koseki_endoscope_sinus_manipulator}. The originality of this non-ferromagnetic robot was its planar parallel architecture, for use in an MRI for real-time navigation. Contrary to a serial architecture, the actuators are all in the same plane. With 2 DOF, the endoscope can tilt during rotation through a parallelogram, with the actuators away from the end-effector. However, the system was bulky and did not allow the surgeon to access the patient's head and operate at the same time.  

In 2006, Nathan et al. developed the AESOP robot, which stands for Automated Endoscopic System for Optimal Positioning robot. This system is a serial 7 DOF robotic arm. Although the architecture is conventional, the originality of the product is the voice control. 
The robot has been evaluated in pre-clinical studies on ten bodies with trans-sphenoidal approaches. It was possible to save certain positions of the endoscope. For example, by saving position 1 in front of the sphenoidal meatus (``Save One''), the surgeon can return to this position at any time by saying ``Return One''. 
However, the robot only responds to simple commands and does not tolerate complex combinations of movements such as those performed manually, and the planning of the movement from the current endoscope position to the saved position does not include a collision test.

In 2011, Fisher et al. studied a robotic system dedicated to endonasal surgery \cite{fischer_endonasal_2011}. 
They designed  a robot attached to the operating table, with a parallel architecture (dual 5-bar linkage allowing 4 DOF),  with an additional DOF for the axial translation of the endoscope. Two joysticks were used for the control, one for linear movements, and one for rotation and tilting. 
But, as Lombard \cite{rapport_SFORL}, the system lacked sufficient workspace, because the tilt and swivel angle was limited to $5^{\circ}$ and the translation to 50mm maximum. 
Indeed, a study by Trévillot et al \cite{trevillot_innovative_2013}, covering 13 endonasal operations, showed that the endoscope (i) tilts from $26^{\circ}$ to $66^{\circ}$, (ii) pivots from $42^{\circ}$ to $71^{\circ}$ and (iii) translates from 70 to 100 mm.  

Lombard et al. presented SurgiDelta \cite{rapport_SFORL}, comprising a 5-DOF passive arm followed by hybrid kinematics consisting of a delta robot (3 translations) followed by 2 revolute joints. Authors reported 52 surgeries in operating room with this robot, which was said to be easy to manipulate and efficiently controlled by voice commands. But no results have been published for now, and no CE marking for marketing. 

\begin{figure} [ht]
\centering
\includegraphics[width=0.3\columnwidth]{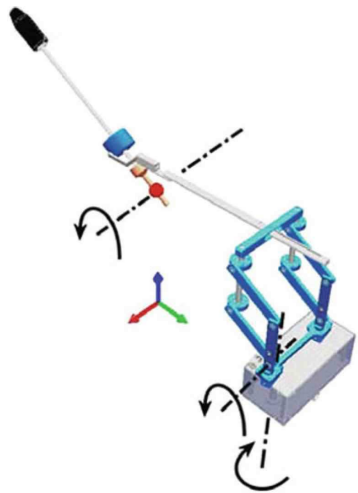}
\caption{EVOLAP architecture, originally  designed for laparoscopic surgery \cite{trevillot_innovative_2013}.}
\label{fig:evolap}
\end{figure}

Another French team \cite{trevillot_innovative_2013} has published a comparison of three architectures for holding an endoscope: (i) EVOLAP, designed for laparoscopic surgery with an RCM generated by 3 parallelograms (Figure \ref{fig:evolap}), (ii) VIPER, a 6-DOF industrial robot with a serial architecture and (iii) HYBRID, a combination of the two previous robots, where the VIPER robot controls the final stage of the EVOLAP system. 
The HYBRID robot was the most suitable for endonasal surgery but, unfortunately, it was only tested in the laboratory.

\begin{figure} [ht]
    \centering
    \includegraphics[width= 0.4\columnwidth]{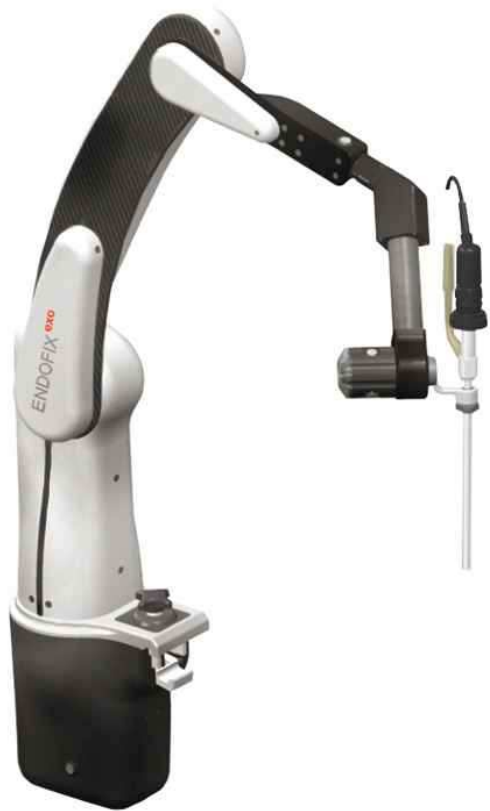}
    \caption{Endofix Exo, robotic endoscope holder, commercialized by AktorMed GmbH, Germany.}
    \label{fig:endofix_exo}
\end{figure}

The Endofix Exo, presented in the section \ref{endos_oreille}, has been studied for endonasal surgery \cite{endofix_exo}, although it was not originally designed for this use. It is a serial architecture with 3 DOFs and fluidic actuators, controlled by a joystick, illustrated Figure \ref{fig:endofix_exo}.
At a pre-clinical stage, 27 endoscopic sinus examinations were performed in the anatomy laboratory. The authors succeeded in reaching all anatomical sites, including the frontal sinus which is more difficult to access. The authors modified the initial prototype designed for abdominal surgery by adding 2 DOFs for a total of 5 DOFs. This modification was necessary due to the absence of a fixed pivot point (trocar).

This device has a CE marking and is marketed for endonasal surgery. On the basis of these data, it fulfils the first objective of releasing the surgeon's hand, but the control is done with a joystick, which partially requires the surgeon's hand. The second objective is to work in collaboration, because the device does not automatically follow the surgeon's gesture. There is no navigation system either.

More recently, a team from the University of Hong Kong has developed a custom-made manipulator for endonasal surgery. 
The innovative feature is the control of the system by the surgeon's foot, hence the name: FREE Foot-controlled Robotic-Enabled Endoscope holder \cite{lin_endoscope_endonasal}. 
The serial architecture has 4-DOF and the foot control is performed by a gyroscopic unit attached to the shoe, which measures the orientation of the foot in real time. By inverting or everting the foot, you choose the joint you wish to move. Then, by moving the heel to the left or right, the pre-selected joint is moved. 
 
This device has been tested for cadaveric dissections \cite{chan_lin_endoscope_endonasal_preclinical}, and the authors stated that the learning time was short, with no problems to perform various surgeries in the sinuses. At this preclinical level, this robot could achieve most of the goals initially cited.

Another system named BEAR (Brescia Endoscope Assistant Robotic Holder) was presented for the endonasal \cite{BEAR}. 
At the preclinical level, this prototype combines an industrial robot to hold the endoscope, and control using the positioning of the surgeon's head (Microsoft Kinect 2.0 sensors). However, BEAR shares with other prototypes the typical limitations of an industrial robot: non-optimal dimensions and excessive inertia.

Friedrich et al. presented a system in 2017 using the positioning arm of the Medineering company \cite{medineering_proof_concept}. This arm has been CE marking for positioning passive adapters and robots in ear, nose and throat surgery, neurosurgery and spinal surgery. It is a passive serial robot with 7 DOF whose joints can be locked in any position. The joints of the holding arm can be released by touch pads on each segment at any time during the procedure. At its distal end, a compact robotic hand with five actuated DOF performs the movement of the endoscope. It is attached directly to the operating table, and its movements are controlled by a customised foot pedal with joystick during the surgery.

\begin{figure} [ht]
   \centering
    \includegraphics[width=0.8\columnwidth]{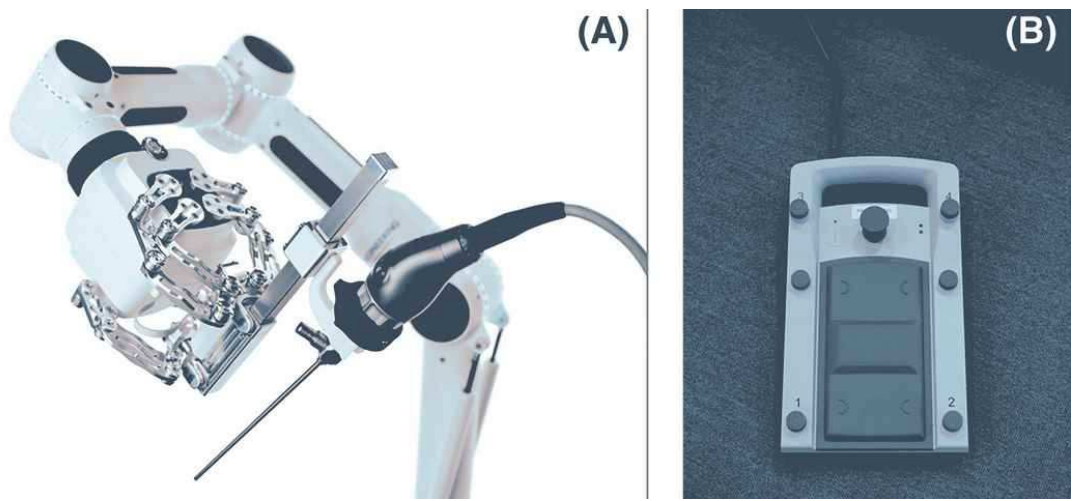}
    \caption{Endoscope Robot (Medineering). A. Overview of the endoscopic positioning system. B. Foot paddle used for steering the endoscopic positioning system \cite{medineering_dcrt}.}
    \label{fig:medineering}
\end{figure}

This robot, shown in Figure \ref{fig:medineering}, is called Endoscope Robot. This robot  has been pre-clinically tested for endonasal surgery \cite{medineering_proof_concept} and has allowed a good visualisation of all sinuses. This led in 2019 to clinical studies for eight endoscopic orbital decompressions \cite{robot_endonasal_medineering_mattheis}. 

The authors state that the installation lasted less than ten minutes and that no complications were reported in 2020 for two stenoses of the nasolacrimal ducts \cite{medineering_dcrt}.
The authors highlighted the need for an irrigation system to clean the endoscope, in order to avoid complete removal of the robot when it is soiled \cite{medineering_proof_concept, medineering_dcrt}.

Up to now, only Endofix Exo and  Endoscope Robot have reached the clinical stage. Furthermore, only the  Endoscope Robot seems to be able to manipulate the endoscope and be controlled without the hand.  However, the robot does not automatically follow the surgeon's movements and does not achieve the navigational objective.
\subsection{Manipulators robots}
In the field of endonasal surgery, some robots have been designed to perform a specific task instead of the surgeon.

In 2003, Wurm \cite{wurm_sinus_endoscope} proposed an automated robot to access the sphenoidal sinus. A preoperative scan was performed to plan the desired trajectory. A tool combined an endoscope and three channels to drill, aspirate and clean. This tool was connected to a Mitsubishi RV-1a industrial robot. This device was only tested in the laboratory.

Yoon et al. \cite{yoon_maxillary_biopsy} presented a bendable endoscope held by a 3 DOF mechanism for maxillary sinus surgery. The endoscope and biopsy tool were controlled by two joysticks. However, this system was only tested on a resin model. One may question the clinical demand for a robot specifically dedicated to maxillary sinus surgery, which is technically the most accessible. 

Burgner et al. \cite{burgner_endonasal_2011} have designed a two-handed remote-controlled robot for endonasal surgery. This robot with concentric tubes provides a central working channel, with a flexible gripper inserted inside. The haptic control was performed using two Phantom Omni master devices. A conventional endoscope was attached to a passive arm. This system can reach certain hard-to-reach areas, thanks to the concentric tube technology. For the time being, it has been tested in an anatomy laboratory for sellar surgery. 

At this time, no manipulator robot has reached the clinical stage in endonasal surgery, maybe due to a low clinical demand, in comparison to endoscopic holders.

\section{Discussion}
Table \ref {tableau_robot_otologie} summarizes the otological robotic systems and Table \ref{tableau_robot_endonasal} summarizes the endonasal robotic systems.

\begin{table} [h!]
 \centering
 \begin{tabularx}{0.8\textwidth} {|p{3cm}|p{1cm}|p{4cm}|p{5cm}|}
 \hline
 \textbf{Robot} & \textbf{Year} & \textbf{Application} & \textbf{Technology}  \\  \hline
 Kuka Kr3\cite{hussong_conception_2008} & 2008 & Drilling robot & Industrial serial robot   \\\hline
 Otobot \cite{danilchenko_otobot_2011} & 2011 & Drilling robot & Industrial robot and custom-built end-effector  \\ \hline
 Vanderbilt Robot \cite{dillon_compact_2015} & 2015 & Drilling robot & 4 DOF serial robot attached to the bone  \\ \hline
 Microtable \cite{kratchman_design_2011} & 2011 & Drilling robot, Microscopic gesture robot (cochlear implant insertion) & Bone attached parallel robot  \\ \hline
 RoboJig \cite{kobler} & 2015 & Drilling robot, Microscopic gesture robot (cochlear implant insertion) & Bone attached parallel robot  \\ \hline
 Smart Micro-Drill \cite{coulson_drill} & 2011 & Drilling robot & Detect the passage of the drill from bone to soft tissue  \\ \hline
 \underline{Hearo} \cite{bell_self-developed_2012} & 2012 & Drilling robot & 5 DOF serial robot, facial nerve monitoring   \\ \hline
 Steady-Hand \cite{steady_hand} & 2004 & Microscopic gesture robot & 7 DOF parallel robot   \\ \hline
 MMTS \cite{inproceedings} & 2014 & Microscopic gesture robot & 6 DOF serial arm, joystick console  \\ \hline
 \underline{Robotol} \cite{miroir_robotol:_2010} & 2010 & Microscopic gesture robot, Endoscopic holding robot & 6 DOF serial robot, space mouse  \\ \hline
\end{tabularx}
\caption{Summary of Otological robotic systems \textit{(underlined robots have reached the clinical stage)}.}
\label{tableau_robot_otologie}
\end{table}

\begin{table} [ht]
 \centering
 \begin{tabularx}{0.8\textwidth} {|p{3cm}|p{1cm}|p{4cm}|p{5cm}|}
 
 \hline
 \textbf{Robot} & \textbf{Year} & \textbf{Application} & \textbf{Technology} \\
 
 \hline
 Robot Koseki et al. \cite{koseki_endoscope_sinus_manipulator} & 2002 & Endoscopic holding robot & planar parallel robot  \\
 
 \hline
AESOP \cite{aesop} & 2006 & Endoscopic holding robot & 7 DOF serial robot, voice control \\

\hline
Fisher et al. Robot \cite{fischer_endonasal_2011} & 2011 & Endoscopic holding robot & 4 DOF parallel robot \\

\hline
SurgiDelta \cite{rapport_SFORL} & 2017 & Endoscopic holding robot & 5 DOF passive arm + hybrid kinematics    \\

\hline
HYBRID \cite{trevillot_innovative_2013} & 2013 & Endoscopic holding robot & 6 DOF serial robot + RCM  \\

\hline
\underline{Endofix Exo} \cite{endofix_exo} & 2015 & Endoscopic holding robot & 5 DOF serial robot  \\

\hline
FREE \cite{lin_endoscope_endonasal} & 2015 & Endoscopic holding robot & 4 DOF serial robot, foot control  \\

\hline
BEAR \cite{BEAR} & 2017 & Endoscopic holding robot & industrial robot, Kinect head control \\

\hline
\underline{Endoscope Robot} \cite{medineering_proof_concept} & 2017 & Endoscopic holding robot & 7 DOF serial robot   \\
 
\hline
Wurm et al. Robot \cite{wurm_sinus_endoscope} & 2003 & Manipulator robot & industrial serial robot \\

\hline
Yoon et al. Robot \cite{yoon_maxillary_biopsy} & 2013 & Manipulator robot & 3 DOF serial robot    \\

\hline
Burgner et al. Robot \cite{burgner_endonasal_2011} & 2011 & Manipulator robot & concentric tubes, haptic control  \\

\hline

\end{tabularx}
\caption{Summary of Endonasal robotic systems \textit{(underlined robots have reached the clinical stage)}.}
\label{tableau_robot_endonasal}
\end{table}

In otolaryngology, surgeons use the Da Vinci robot to treat certain cancers. However, this robot cannot be used in ear or sinus surgery because of its size. Some of the robots presented in this review have been designed to assist the surgeon in ear or sinus surgery, for several purposes.

In ear surgery, drilling robots  are mainly dedicated to cochlear implant surgery. They secure the path to the round window, close to the facial nerve. Only one robot, the HEARO system, has recently obtained the CE mark, and has been able to secure this procedure to access the cochlea. However, this robot is specifically dedicated to this particular surgery, and cannot be used in other otological surgeries, and even less in sinus surgery. Conversely, Robotol is a robot capable of performing a precise gesture.
It can be useful in cochlear implant surgery, not to access the cochlea like the HEARO robot, but to gently insert the implant. The Robotol can also be used in other otologic surgeries, such as ossicular replacement or endoscopic procedures. This versatility could help justify its purchase by a hospital. However, this robot does not eliminate manual control of its mobility and is not suitable for sinus surgery. No other otological robot has reached the clinical stage at this time. However, the high number of robots studied tends to demonstrate the high expectations of surgeons in this field.

In sinus surgery, most robots have been designed to hold an endoscope. Indeed, most surgeries have been performed by endonasal approach for about twenty years, for sinus surgery (infections, polyps, ...) and skull base surgery (cancer mainly). Several robots are designed to perform a specific task in place of the surgeon, but none of them have reached the clinical stage.  Among the endoscopic holding robots, only the Endofix Exo and the Endoscope Robot are marketed. The Endofix Exo has proven its ability to reach all anatomical sites during endonasal procedures. However, this robot remains manually controlled.  The Endoscope Robot completes the objective of being controlled without the hand, thanks to a foot pedal with joystick. However, none of these robots automatically follows the surgeon's movements, nor do they include a navigation system. 

Many prototypes have been studied, but few have reached the clinical stage. It seems that in recent years, teams have tested their devices at the pre-clinical stage, and sometimes obtained a CE mark to begin clinical use.

Up to now, no robot is suitable for ear and sinus surgery. However, ear and sinus surgery does have some common features: it is performed in the same ENT (Ear, Nose and Throat) speciality, on the same scale of workspace and with endoscopes of the same size. It would be interesting to share the same robot for all endoscopic procedures in ENT surgery. 

In addition, like all innovative processes, these robots must now demonstrate their benefit in clinical use. 

\section{Conclusions}
In this review, we have established a classification of robots for ear surgery and sinuso-nasal surgery, according to their clinical functions.  
Many robots are being studied, with different approaches to clinical applications, control and kinematics, but very few robots have reached the clinical level by obtaining a CE marking.

For ear surgery, most robots are dedicated to cochlear implant surgery. But only the Robotol, with an extra arm, is able to assist the surgeon during endoscopic ear surgery.

Conversely, for endonasal surgery, most robots are designed to hold the endoscope. Two have now reached the clinical stage: Endofix Exo and Endoscope Robot. However, only the Robot Endoscope offers control with one foot, freeing the surgeon's hand even when controlling the endoscope. 

At the moment, no endoscopic robot crosses the border between ear surgery and endonasal surgery. It would be interesting to share the same robot for all endoscopic procedures in otolaryngologic surgery, even in neurosurgery, to avoid, in the future, a hospital multiplying purchases of robots with associated training. Another means of improvement is freehand control of robots, and navigation with augmented reality to make surgery safer. 
\begin{acknowledgment}
First author received the grant from ``Association Française d'Oto-Neurochirurgie''. The project is receiving financial support from the NExT (Nantes Excellence Trajectory for Health and Engineering) Initiative and the Human Factors for Medical Technologies (FAME) research cluster.
\end{acknowledgment}
%
\bibliographystyle{asmems4}
\bibliography{asme2e}
\cleardoublepage
\section{Figure captions}
Fig. 1: The comparison of the number of instruments possible to use simultaneously while using an endoscope and a microscope.

Sub-captions:

(a) Surgeon using 2 instruments with a microscope

(b) Use of 1 hand to hold endoscope limits the instruments

Fig. 2: Robotol system, initially designed for microscopic gesture. Later, a second arm has been added to hold an endoscope. Source:
Collin Medical.

Sub-captions: 

(a) Serial architecture of the Robotol

(b) Robotol in its commercialised version with CE marking.

Fig. 3: EVOLAP architecture, originally designed for laparoscopic surgery [29].

Fig. 4: Endofix Exo, robotic endoscope holder, commercialized by AktorMed GmbH, Germany.

Fig. 5: Endoscope Robot (Medineering). A. Overview of the endoscopic positioning system. B. Foot paddle used for steering the
endoscopic positioning system [35].

\cleardoublepage
\section{Table captions}

Table 1: Summary of Otological robotic systems (underlined robots have reached the clinical stage).

Table 2: Summary of Endonasal robotic systems (underlined robots have reached the clinical stage).
\end{document}